\documentclass[10pt,twocolumn,letterpaper]{article}

\usepackage[pagenumbers]{cvpr} 

\usepackage{graphicx}
\usepackage{amsmath}
\usepackage{amssymb}
\usepackage{booktabs}
\usepackage{xcolor}
\usepackage{soul}
\usepackage{caption}

\usepackage[pagebackref,breaklinks,colorlinks]{hyperref}

\usepackage[capitalize]{cleveref}
\crefname{section}{Sec.}{Secs.}
\Crefname{section}{Section}{Sections}
\Crefname{table}{Table}{Tables}
\crefname{table}{Tab.}{Tabs.}

\def\cvprPaperID{11369} 
\def\confName{CVPR}
\def\confYear{2023}

\begin{document}

\title{A Geometric Model for Polarization Imaging on Projective Cameras}

\author{Mara Pistellato \hspace{2cm}Filippo Bergamasco\\
DAIS - Ca'Foscari University of Venice\\
Via Torino 155, 30172 Venice - Italy\\
{\tt\small \{mara.pistellato, filippo.bergamasco\}@unive.it}
}
\maketitle

\begin{abstract}
The vast majority of Shape-from-Polarization (SfP) methods work under the oversimplified assumption of using orthographic cameras. Indeed, it is still not well understood how to project the Stokes vectors when the incoming rays are not orthogonal to the image plane.
We try to answer this question presenting a geometric model describing how a general projective camera captures the light polarization state. Based on the optical properties of a tilted polarizer, our model is implemented as a pre-processing operation acting on raw images, followed by a per-pixel rotation of the reconstructed normal field. In this way, all the existing SfP methods assuming orthographic cameras can behave like they were designed for projective ones. Moreover, our model is consistent with state-of-the-art forward and inverse renderers (like Mitsuba3 and ART), intrinsically enforces physical constraints among the captured channels, and handles demosaicing of DoFP sensors.
Experiments on existing and new datasets demonstrate the accuracy of the model when applied to commercially available polarimetric cameras.\vspace{-0.4cm}
\end{abstract}

\section{Introduction}
\label{sec:intro}

When light reflects on a surface, its polarization changes according to
well-established physical rules that describe such interaction.  Among all the
factors involved in the process, the surface properties and its orientation
also determine the final polarization state of the captured beam.  For this
reason the literature counts several works aiming at recovering surface
properties from images taken with a rotating linear polarizer in front of the
camera
\cite{wolff1993constraining,atkinson2006recovery,miyazaki2003polarization}.

Recently, the availability of Division-of-Focal-Plane (DoFP) cameras raised the
interest of the Computer Vision community in polarization-related applications.
In particular, light polarization is typically exploited to recover surface
orientation in the so-called Shape from Polarization (SfP) applications.  Such
methods are designed to recover the 3D shape of acquired objects from a single
view thanks to the information intrinsically encoded in the polarimetric images
\cite{miyazaki2003polarization,shakeri2021polarimetric,baek2018simultaneous,taamazyan2016shape}.
Several approaches in the SfP domain combine polarimetric imaging with other
cues coming from classical techniques such as stereo
\cite{fukao2021polarimetric}, multi-view
\cite{atkinson2005multi,cui2017polarimetric,zhao2020polarimetric,chen2018polarimetric}
or shading \cite{ngo2015shape,smith2018height}.  Moreover, some works presented
in the recent past propose data-driven approaches to perform shape from
polarization \cite{ba2020deep,lei2022shape} exploiting specially-made datasets
which involve the use of polarimetric cameras and 3D scanners capturing a wide
range of subjects, from small objects to entire buildings.  Other practical
applications of polarimetric data involve pose estimation
\cite{cui2019polarimetric}, material classification
\cite{tominaga2008polarization} and HDR reconstruction \cite{wu2020hdr}, while
some methods propose to extend the dense SLAM system
\cite{yang2018polarimetric}.  Finally, some works propose a recovery of both
surface geometry and refractive index via multispectral polarimetric imaging
\cite{huynh2013shape}.

Despite the undeniable contribution of these approaches in the present-day
relevant literature, almost all of them rely on the quite unrealistic
assumption of operating with an orthographic camera.  Indeed, the basic
equations often presented in such works are designed for a model where light
rays hit the sensor perpendicularly: this is pointed out in some early papers
such as \cite{rahmann2001reconstruction,rahmann2000polarization}.  The majority
of proposed approaches employ pinhole cameras, in which light beams hit the
image plane with an angle that depends on the camera geometry.  In practice,
assuming an orthographic model while acquiring with a projective camera leads
to non-negligible errors and deformations, especially when we are interested on
areas near the image borders or if we have short focal lengths.  The authors of
\cite{chen2022perspective,lei2022shape} highlight the described problem and try
to formulate a solution for the perspective deformation and the representation
of polarization data in non-orthographic imaging devices.

\subsection{Related Works}

In the literature we find an abundance of methods focusing on Shape from
Polarization but very few works trying to understand how to deal with
projective cameras. Indeed, we are concerned that almost all the methods assume
(explicitly or even implicitly) to have an orthographic camera, without
considering how much this assumption might affects the results.
Only recently, two works addressed this problem by providing two different
solutions. Chen et al.~\cite{chen2022perspective} gave the first purely
geometric relationship between the polarization phase angle and the azimuth
angle of the observed surface normal. Albeit interesting, the model has two
limitations. First, it does not account for normal elevation so its
applicability is limited to some specific contexts (i.e. single-view recovery
of planar surfaces or multi-view estimation of normals). Second, it does not
provide a direct description of how the full polarization state (i.e. the
Stokes vector) is captured, but only how the Angle of Linear Polarization is
affected by the ray direction. The second work, proposed by Lei et.
al~\cite{lei2022shape}, is based on the observation that ``the polarization
representation is highly influenced by the viewing direction". Their solution
consists on a Convolutional Neural Network taking in input the captured image
together of a viewing encoding providing cues to the camera intrinsic
properties. Accuracy of the resulting normal maps are currently unmatched, but
such data-driven approaches give no explicit information on how the model works
internally. As often happens with pure learning-based solutions, the resulting
model is a black-box which can hardly be generalized to different contexts.

\subsection{Our contributions}

In this paper we give the first complete mathematical description of how the
scene polarization is captured with a projective camera. Our model is based on
the optical properties of the tilted polarizer~\cite{korger2013polarization}
and formulated by applying the algebra of Mueller matrices. In this respect,
the main contribution of our work is that the analysis of what happens in a
projective camera not based on pure empirical evidence (i.e. data-driven) but
on the physical well-studied background of how light polarization is
represented and transformed when interacting with the camera optical elements.
In doing so, we show how some formulas commonly used in SfP simply derive by
the application of particular Mueller matrices to the incoming Stokes vectors.

One of the key advantages of the resulting model is that it can be applied as a
pre-processing on the acquired image and a post-processing of the estimated
normal field. Therefore, it can be used seamlessly on existing SfP methods
designed with the orthographic assumption in mind. We adopted the same
conventions assumed in popular direct and inverse renderers believing that this
will simplify the creation of synthetic datasets closely simulating what can be
captured with a real polarimetric camera. This topic is of a pivotal importance
to train state-of-the-art deep learning based models and for testing existing
methods against a controllable Ground Truth.  Finally, the model is equally
valid for DoFP and Division-of-Time (DoT) cameras, and generalizes to any
number of linear polarizers involved in the acquisition process.

\newpage \section{Preliminaries} To understand the theoretical background of
the proposed model, we briefly summarize some basic notions about light
polarization. The goal here is to highlight crucial aspects that are sometimes
neglected when approaching shape-from-polarization. Please refer to
\cite{collett2005field,bass2009handbook} to learn more about these concepts.

\subsection{Light polarization}

Any visible light ray consists of two orthogonal electric field components
$E_x, E_y$ oscillating in the plane transverse to the direction of propagation
$\vec{k}$. Without loss of generality, $\vec{k}$ can be set coincident with the
$z$-axis, so that $E_x, E_y, z$ form an orthogonal reference system in which:
\begin{eqnarray}\label{eq:ellipse} E_x(z,t) &=& A_x \cos\big(\omega t -
\frac{2\pi}{\lambda} z + \delta_x\big)\nonumber \\ E_y(z,t) &=& A_y
\cos\big(\omega t - \frac{2\pi}{\lambda} z + \delta_y\big).  \end{eqnarray}
        \noindent $A_x$ and $A_y$ are two wave amplitudes, $\omega$ is the
        angular frequency, $\lambda$ is the wavelength and $\delta_x, \delta_y$
        are two arbitrary phases. The point $\begin{pmatrix} E_x(z,t) &
        E_y(z,t)\end{pmatrix}$ traces a so called \emph{polarization ellipse}
        when discarding the time-space propagator $\omega t -
        \frac{2\pi}{\lambda}z$ (i.e. when observing the two waves ``projected"
        in the $x-y$ plane). Orientation angle and eccentricity of such ellipse
        describe the polarization state of light. For example, when $A_y=0,
        A_x>0$ the optical field oscillates horizontally and we have a
        \emph{linearly horizontal polarized light} (polarization ellipse
        degenerates to an horizontal segment). When $A_x=A_y,
        \delta_x-\delta_y=0$ we have polarized light with an \emph{Angle of
        Linear Polarization} (AoLP) of $45^{\circ}$.

An important thing has to be noted here. When we discuss about polarization
angles (horizontal, $45^\circ$, etc.) we have to make clear the orthogonal
reference system in which the electric field components are expressed. The
third axis is implicitly known, since we \emph{always assume the $z$-axis being
the direction of propagation}. The $x$-axis can be any unitary vector
orthogonal to $z$, thus providing a reference to which such angles are
expressed. Since the system is orthogonal, the $y$-axis is uniquely determined
as $y=x \times z$. Therefore, a light ray with an AoLP of $45^\circ$ can be
seen as a ray with an AoLP of $0^\circ$ if we rotate the reference system
$45^\circ$ counter-clockwise around the $z$-axis. 

\subsection{Stokes parameters and Mueller matrices}

$E_x, E_y$ are not directly measurable, so a different formulation is usually
preferred. Taking a time average of Eqs. \ref{eq:ellipse} yield the definition
of the four quantities:
            \begin{eqnarray} S_0 &=& A_x^2 + A_y^2 \nonumber \\ S_1 &=& A_x^2 -
            A_y^2 \nonumber \\ S_2 &=& 2 A_x A_y \cos( \delta_y - \delta_x)
        \nonumber\\ S_3 &=& 2 A_x A_y \sin( \delta_y - \delta_x) \end{eqnarray}

\noindent called \emph{Stokes polarization parameters}, usually grouped in the
Stokes vector $S=\begin{pmatrix}S_0 & S_1 & S_2 &
S_3\end{pmatrix}^T$. This formulation is powerful because can
describe partially polarized light as a mixture of unpolarized
and completely polarized light:
\begin{equation}\label{eq:mixture}
S=(1-\rho)\begin{pmatrix}S_0\\0\\0\\0\end{pmatrix} + \rho
\begin{pmatrix}S_0\\S_1\\S_2\\S_3\end{pmatrix}
\end{equation}

\noindent where $\rho=\sqrt{S_1+S_2+S_3} / S_0$ is called \emph{Degree of
Linear Polarization} (DoLP). It is easy to observe from Eq.\ref{eq:mixture}
that $S_0$ is the intensity of light and the AoLP $\phi$ is given by
$\phi=\mbox{atan2}(S_2,S_1)$.

Stokes parameters can be measured by letting the light rays pass trough special
materials called polarizers and retarders. Such elements transform the input
Stoke vector by means of a linear transformation described by a $4 \times 4$
Mueller matrix $\mathbf{M}$. For example, the Mueller matrix of an ideal linear
polarizer with transmission axis oriented with an angle $\alpha$ with respect
to the $x$-axis is:

\begin{equation}\label{eq:linearpolarizer}
    \mathbf{M_\alpha}=\frac{1}{2}\begin{pmatrix} 1 & \cos 2\alpha & \sin 2\alpha & 0 \\
    \cos 2\alpha & \cos^2 2\alpha & \sin 2\alpha \cos 2\alpha & 0 \\
    \sin 2\alpha & \sin 2\alpha \cos 2\alpha & \sin^2 2 \alpha & 0 \\
    0 & 0 & 0 & 0 \end{pmatrix}.
\end{equation}

Also in this case, the reference frame matters and cannot be chosen
arbitrarily. The Mueller matrix of a polarizing element must be defined in a
system coincident to the one in which the input/output Stokes vectors are
expressed. If that's not the case, a rotator $R$ (around the $z$-axis) must be
applied to align the reference systems, thus getting:

\begin{equation*}
    S' = R^T \mathbf{M} R S
\end{equation*}

\noindent where $S$ and $S'$ are input and output Stokes vectors respectively.
In any case, the $z$-axis of $S, S'$ and $\mathbf{M}$ is fixed to the direction
in which the ray travels. We stress the fact that Stokes vectors and Muller
matrices alone are meaningless without an associated reference frame.

\subsection{Polarimetric cameras}

A polarimetric camera can measure the first $3$ components of a Stokes
vector\footnote{Circular polarization is relatively rare in nature
\cite{cronin2011patterns} and therefore it is usually not accounted for.} using
an array of linear polarizers placed in front of a standard CCD or CMOS sensor.
This can be implemented by adding fixed filters directly onto the pixel-grid
(DoFP cameras), or by taking multiple pictures while rotating a linear polarizer
in front of the lenses (DoT cameras). In both the cases, the polarizers are
parallel to the image plane so that the \emph{transmitting axis} is a vector
$\vec{t}_\alpha=\begin{pmatrix}\cos \alpha & \sin \alpha & 0\end{pmatrix}^T$
expressed in the camera reference frame.

Regardless the intrinsic properties of the camera (Orthographic or Projective),
for each pixel a set of $\mathcal{I}=\{ I_{\alpha_0}, I_{\alpha_1}, \ldots,
I_{\alpha_N} \}$ intensities are captured by using linear polarizers with angles
$\mathcal{A}=\{\alpha_0, \alpha_1, \ldots \alpha_N\}$ respectively. The task is
to recover the incoming Stokes vector $S=\begin{pmatrix}S_0 & S_1 & S_2
& 0\end{pmatrix}^T$ from those observations.

\subsection{Orthographic model}

\begin{figure*}
    \centering
    \includegraphics[width=\linewidth]{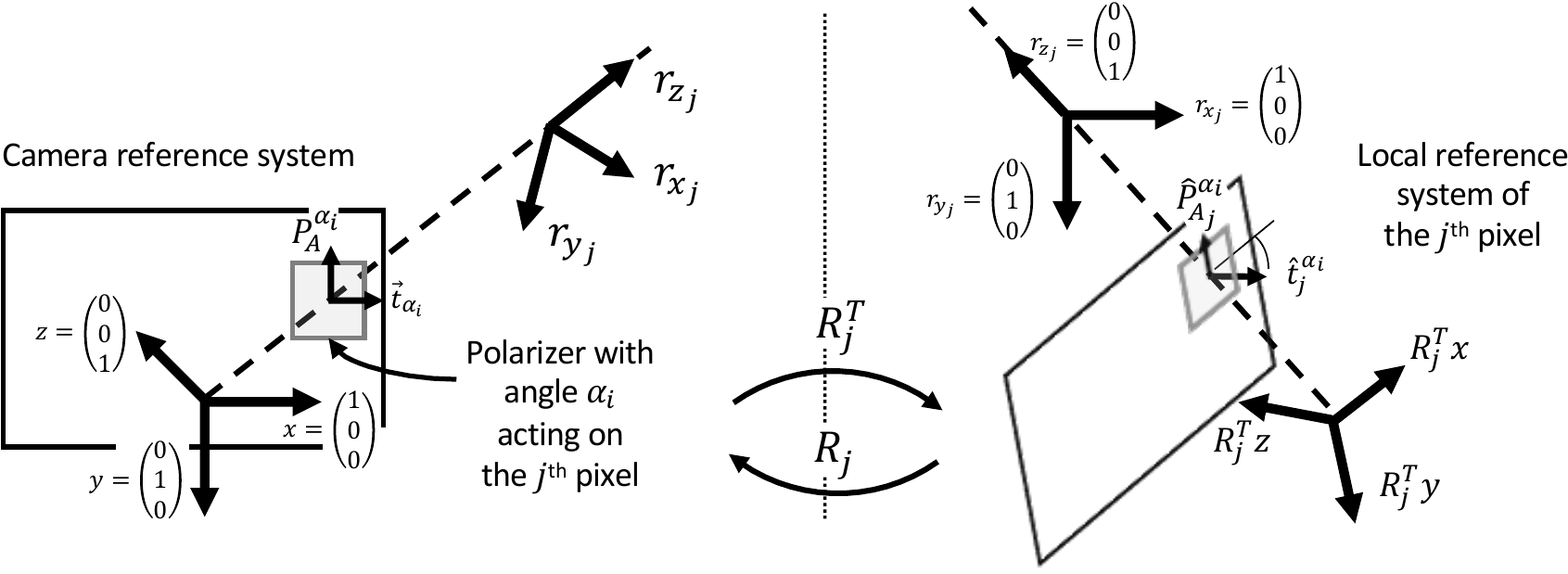}
    \caption{Sketch of the proposed model. In the camera reference system (left), the ray passing through the $j^{th}$ pixel (dashed line) is not propagating parallel to the $z$-axis. Therefore, the effect of a polarizer with angle $\alpha_i$ placed parallel to the image plane cannot be described with formulas used for orthographic cameras. To solve this, we define a local reference system for each ray (right) oriented along the direction of propagation. Since the linear polarizer is tilted, its effective angle $\hat{\alpha}_i^j$ is different from the actual orientation it has on the sensor. The tilted polarizer model \cite{korger2013polarization} is used to compensate this distortion and recover the correct Stokes vector $S_j$ in the reference frame of the $j^{th}$ pixel. Finally, a rotation $R_j$ allow us to map normal vectors estimated from $S_j$ back to the camera reference system.}
    \label{fig:model}
\end{figure*}

The orthographic camera model is simple to deal with because all the rays are
assumed to enter perpendicularly to the image plane. Therefore, we can
conveniently set the reference frame of all the rays such that the first two
axes follows the horizontal and vertical ordering of the pixels, and the
$z$-axis coincides with the camera optical axis (i.e. the direction in which all
the rays are propagating). Note that, since the $y$-axis is commonly oriented
downward (following the top-down ordering of image pixels) and the $x$-axes
rightward, the polarization angles are measured clockwise from the $x$-axis
instead of the classical counter-clockwise notation. So, care must be taken when
using formulas involving polarization angles.

Since $S_0$ represents the light intensity, for each pixel we can easily relate
the intensities $\mathcal{I}$ to the incoming Stokes vector $S$ by writing a set
of linear equations:
\begin{equation}\label{eq:linearsystem} \begin{pmatrix} I_{\alpha_0} \\
I_{\alpha_1}\\ \vdots \\ I_{\alpha_N}\end{pmatrix} = \begin{pmatrix}
\mathbf{M}_{\alpha_0}^1 \\ \mathbf{M}_{\alpha_1}^1 \\ \vdots \\
\mathbf{M}_{\alpha_N}^1  \end{pmatrix} \begin{pmatrix}S_0 \\ S_1 \\ S_2 \\
0 \end{pmatrix} \end{equation}
\noindent where $\mathbf{M}_{\alpha_i}^1$ is the first row of the matrix
$\mathbf{M}_{\alpha_i}$ (defined in  Eq.~\ref{eq:linearpolarizer}). At this
point, $S$ can be estimated by solving Eq.~\ref{eq:linearsystem} in
a least-squares sense \cite{nayar1997separation, huynh2010shape}, usually by
first expressing each equation $I_{\alpha_i}=\mathbf{M}_{\alpha_i}^1 S$ in terms
of AoLP $\phi$ and DoLP $\rho$:
\begin{eqnarray}\label{eq:systemtrigonometric} I_{\alpha_i} &=& \frac{I_{\max}
+ I_{\min}}{2}+\frac{I_{\max} - I_{\min}}{2}\cos( 2\alpha_i - 2\phi)\nonumber\\
\rho &=& \frac{I_{\max} - I_{\min}}{I_{\max} + I_{\min}}. \end{eqnarray}

A special case is given by PFA cameras composed by just four polarizers arranged
with angles $\alpha_{0,1,2,3}=0^\circ, 45^\circ, 90^\circ, 135^\circ$. In this
setting, the trigonometric Eqs.~ \ref{eq:systemtrigonometric} have a simple
closed-form solution:
\begin{eqnarray} S_0 &=& I_0 + I_{90}\nonumber\\ S_1 &=& I_0 - I_{90}\nonumber\\
S_2 &=& I_{45} - I_{135}.\label{eq:PFAstokes} \end{eqnarray}

\noindent Note that, for the orthogonality of such polarizers, intensities are subject to the constraint $I_0+I_{90}=I_{45}+I_{135}$. So, $S_0$ can alternatively be computed as $I_{45}+I_{135}$.

\subsection{The proposed model}

The problem of dealing with a projective camera arise because rays propagate in
different directions and none of them, except the central one, is parallel to
the optical axis. Therefore, we need to: \begin{enumerate} \item Define
a reference system unique for each ray. The $z$-axis must always point to the
direction of propagation but we still have freedom of choice for the other two.
(\cref{sec:referencesystem}) \item Understand what happens to the Stokes vector
when a ray traverse a linear polarizer \emph{tilted} with respect to the ray
direction (\cref{sec:tiltedpolarizer}) \end{enumerate}

\subsection{Local ray reference system}\label{sec:referencesystem}

We suppose to know the matrix $\mathbf{K}$ of intrinsic camera parameters that
can be estimated with any popular calibration tool. For each pixel
$p_j=\begin{pmatrix}u_j & v_j\end{pmatrix}$ in the image plane, the
corresponding \emph{exiting ray} is the 3D unitary vector:
\begin{equation} \vec{r}_{z_j} = \frac{\mathbf{K}^{-1}\begin{pmatrix}u_j & v_j
& 1\end{pmatrix}^T}{\| \mathbf{K}^{-1}\begin{pmatrix}u_j & v_j
& 1\end{pmatrix}^T\| }. \end{equation}

The third axis of $p_j$'s local reference system must be $\vec{r}_{z_j}$ because
it represents its direction of propagation\footnote{To be precise,
$-\vec{r}_{z_j}$ is the true direction but changing the sign will not affect the
orientation and the Mueller calculus still applies.}. For the other two axes we
have freedom of choice. Without loss of generality, we set:
\begin{eqnarray} \vec{r}_{x_j} &=& \frac{\begin{pmatrix}0 & 1 & 0\end{pmatrix}^T
\times \vec{r}_{z_j}}{\| \begin{pmatrix}0 & 1 & 0\end{pmatrix}^T \times
\vec{r}_{z_j} \|} \label{eq:rx} \\ \vec{r}_{y_j} &=& \vec{r}_{z_j} \times
\vec{r}_{x_j} \end{eqnarray}
\noindent thus creating the orthogonal reference system
$\mathcal{P}_j=(\vec{r}_{x_j}, \vec{r}_{y_j}, \vec{r}_{z_j})$ shown in
\cref{fig:model}. Note that there is no physical reason to prefer
$\vec{r}_{x_j}$ as defined in Eq.~\ref{eq:rx} since any other vector orthogonal
to $\vec{r}_{z_j}$ would have been equally valid for the model. However, it
makes sense to set the frame such that its horizontal axis is aligned with the
camera $x$-axis (indeed, $r_{x_j}\perp y$ by construction). Moreover, this is
the same convention used by the state-of-the-art polarization aware renderers
Mitsuba \cite{jakob2022mitsuba3} and ARC \cite{WilkieARC} so synthetically
generated images can be easily compared with real images processed with our
model. Finally, we define the matrix:
\begin{equation}\label{eq:rj} R_j = \begin{pmatrix} \vert & \vert & \vert \\
\vec{r}_{x_j} & \vec{r}_{y_j} & \vec{r}_{z_j} \\ \vert & \vert & \vert
\end{pmatrix} \end{equation}
\noindent mapping vectors from $\mathcal{P}_j$ to the camera reference system.

\subsection{Tilted polarizers}\label{sec:tiltedpolarizer}

Polarimetric cameras are constructed so that the array of polarizers are
parallel to the image plane. Therefore, in each local reference system
$\mathcal{P}_j$, such polarizers are tilted and the transmission axis is in
general not orthogonal to $\vec{r}_{z_j}$. For this reason,
Eq.~\ref{eq:linearsystem} is not correctly defined because each
$\mathbf{M}_{\alpha_i}$ lie in the camera reference frame which is not aligned
with the ray direction of propagation.

To solve the problem, we follow the empirical model of Korger et
al.\cite{korger2013polarization} assuming to have polarizing elements made of
anisotropic absorbing and scattering particles. According to such model, the
\emph{effective transmitting axis} $\mathbf{\hat{t}}_j$ of a tilted polarizer is
orthogonal to both $\vec{r}_{z_j}$ and the \emph{absorbing axis} $\hat{P_A}_j$.
In other words, the effect of a tilted polarizer with angle $\alpha$ is
equivalent of a linear polarizer aligned with the ray direction (so that its
effect on the Stokes vector can it be expressed with a Mueller matrix) but with
a different \emph{effective angle} $\hat{\alpha}$.

We can obtain the Mueller matrix $\mathbf{M}_{\hat{\alpha}_i^j}$ of the $i^{th}$
tilted polarizer on in the local frame of the $j^{th}$ pixel by computing the
effective angle $\hat{\alpha}_i^j$ as follows:

\begin{eqnarray}
    \hat{P}_{A_j}^{\alpha_i} &=& R_j^T \begin{pmatrix} \cos(\alpha_i+\pi/2) \\ \sin(\alpha_i+\pi/2) \\ 0 \end{pmatrix}\label{eq:pa}\\
    \mathbf{\hat{t}}_j^{\alpha_i} &=& \begin{pmatrix}\cos\hat{\alpha}_i^j \\ \sin\hat{\alpha}_i^j \\ 0\end{pmatrix} = \frac{\begin{pmatrix}0 & 0 & 1\end{pmatrix}^T \times \hat{P}_{A_j}^{\alpha_i}}{\|\begin{pmatrix}0 & 0 & 1\end{pmatrix}^T \times \hat{P}_{A_j}^{\alpha_i}\|}.\label{eq:t}
\end{eqnarray}

\noindent Equation \ref{eq:pa} expresses the absorbing axis of the $i^{th}$
polarizer in the local reference frame of the ray. We add $\frac{\pi}{2}$ to the
polarizer angle $\alpha_i$ because we assume the absorbing axis being orthogonal
to the transmission axis. Then, Eq.~\ref{eq:t} computes the effective
transmitting axis $\mathbf{\hat{t}}_j^{\alpha_i}$, which is orthogonal to the
ray direction of propagation by construction. The angle of
$\mathbf{\hat{t}}_j^{\alpha_i}$ in the local reference frame of the ray gives
the effective angle $\hat{\alpha}_i^j$.

To summarize, a perspective camera can be used like an orthographic one with the following precautions: 
\begin{enumerate}
    \item Each recovered Stokes vector is defined on a different reference
    frame, depending to the ray direction for that pixel. Consequently, normal
    vectors computed from the Stokes lie on different frames as well, and must
    be transformed back to the camera reference frame through $R_j$.

    \item Even if the camera uses a few different polarizers with angles in
    $\mathcal{A}$, each pixel will observe equivalent polarizers with
    a different set of angles $\hat{\mathcal{A}}_j=\{\hat{\alpha}_1^j,
    \hat{\alpha}_2^j,\ldots,\hat{\alpha}_N^j\}$. This implies that a linear
    system like the one shown in Eq.~\ref{eq:linearsystem} must be solved in any
    case, since simpler closed form solutions (See Eq.~\ref{eq:PFAstokes})
    cannot be valid simultaneously for all the pixels.
\end{enumerate}

\subsection{How to use our model}\label{sec:howtouse}

The main advantage our model is that it does not require to reformulate
existing SfP methods designed with the orthographic camera assumption. Indeed,
we can synthetize new images that would have been seen with ideal (non tilted)
polarizers rotated at angles $0^\circ,45^\circ,90^\circ,135^\circ$. After this
pre-processing, Eq.~\ref{eq:PFAstokes} can be used get the full Stokes vector,
AoLP and DoLP needed to compute the normal vector field. When normals are
recovered, they will be expressed in the local reference frame of each pixel.
So, every vector must be transformed to the common camera reference frame by
applying the rotation $R_j$ (Eq.~\ref{eq:rj}). This is a post-processing
operation that can be transparently applied to the output of any SfP method. We
now sketch the basic steps to be performed to embed our model in existing or
future approaches for Shape-from-Polarization:
\begin{enumerate}
    \item Calibrate the camera to get the intrinsic matrix $\mathbf{K}$
    \item Compute per-pixel reference systems and transformation matrices $R_j$ using Eq.~\ref{eq:rj}
    \item For each pixel $j$ and for each polarizer angle $\alpha_i$, compute the effective polarizer angle $\hat{\alpha}_i^j$ using Eq.~\ref{eq:t}
    \item Solve the linear system in Eq.~\ref{eq:linearsystem} but using $\mathbf{M}_{\hat{\alpha}_i^j}^1$ instead of $\mathbf{M}_{\alpha_i}^1$ to compensate the effect of the tilted polarizers. This will produce a Stokes vector $S_j$ for each pixel. Note that this vector is not expressed in the camera reference frame but in the local pixel reference frame $\mathcal{P}_j$.
    \item If the SfP method directly accepts the Stokes vector, use the ones computed in the previous step. Otherwise, \textbf{pre-process} the data by synthetizing a new set of images $\hat{\mathcal{I}}=\{\hat{I}_0,\hat{I}_{45},\hat{I}_{90},\hat{I}_{135}\}$  where the image $\hat{I}_\gamma$ is obtained by multiplying each Stokes vector $S_j$ with the Muller matrix $\mathbf{M}_\gamma$.
    \item Since Stokes vector are in local frames, the 3D normal vectors obtained by SfP are expressed in the local reference frames as well. Therefore, whenever a normal $\vec{n}_j$ is estimated for a pixel $j$, it must be transformed back to the camera reference system by computing $R_j \vec{n}_j$. This is essentially a \textbf{post-processing} operation to be applied to the resulting normal field.
\end{enumerate}

\begin{figure*}
    \centering
    \includegraphics[width=0.33\linewidth]{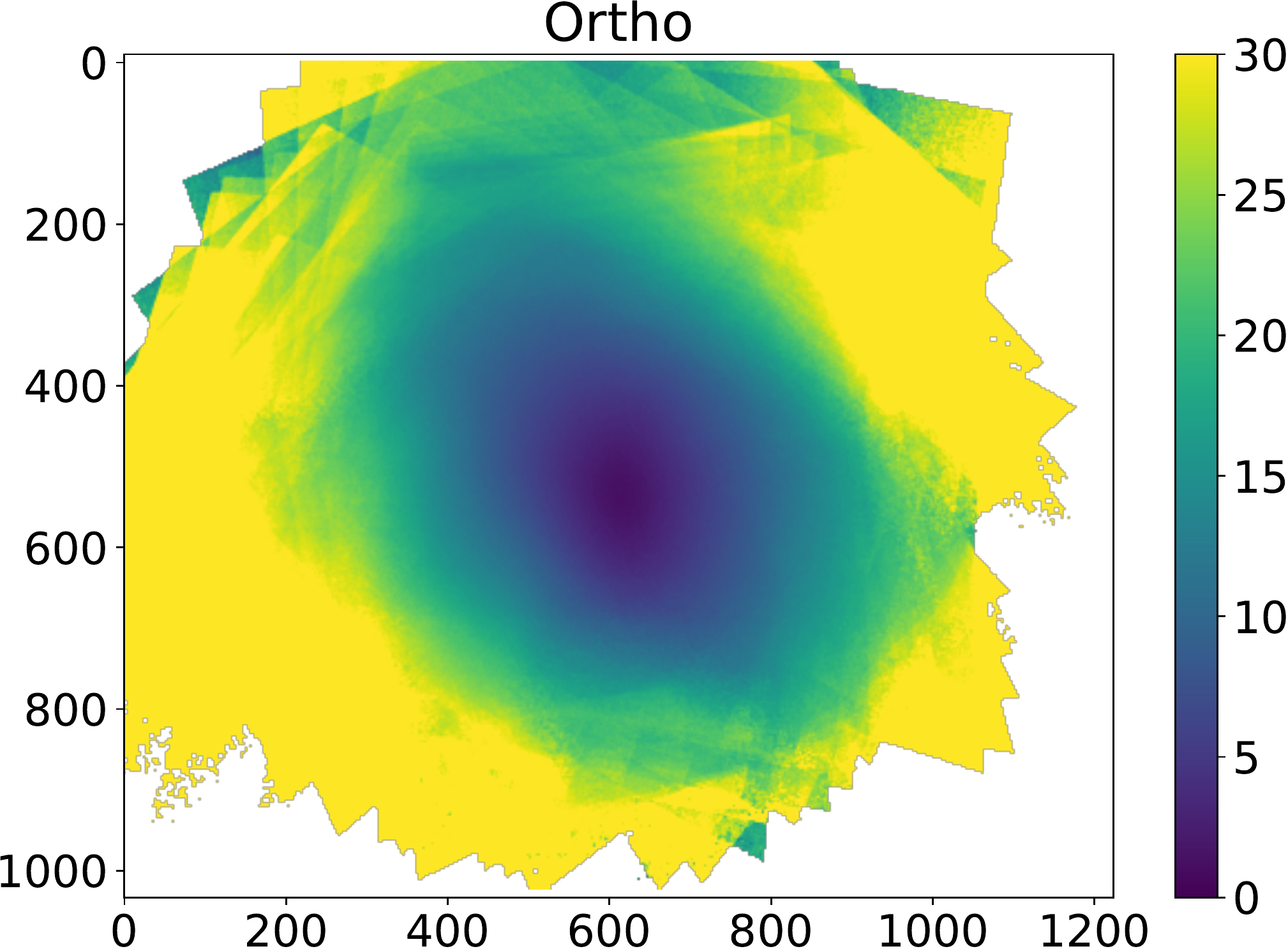}
    \includegraphics[width=0.33\linewidth]{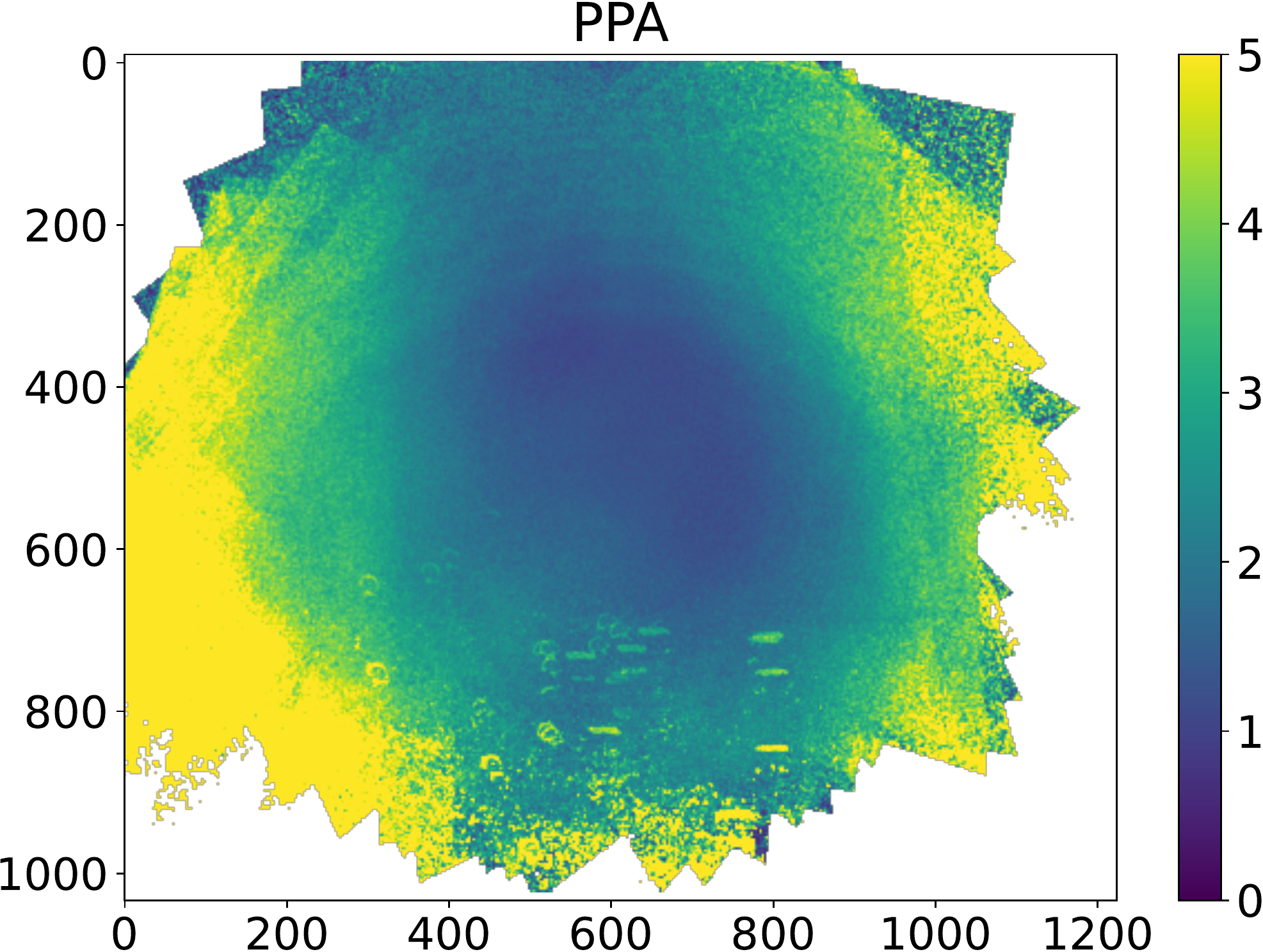}
    \includegraphics[width=0.33\linewidth]{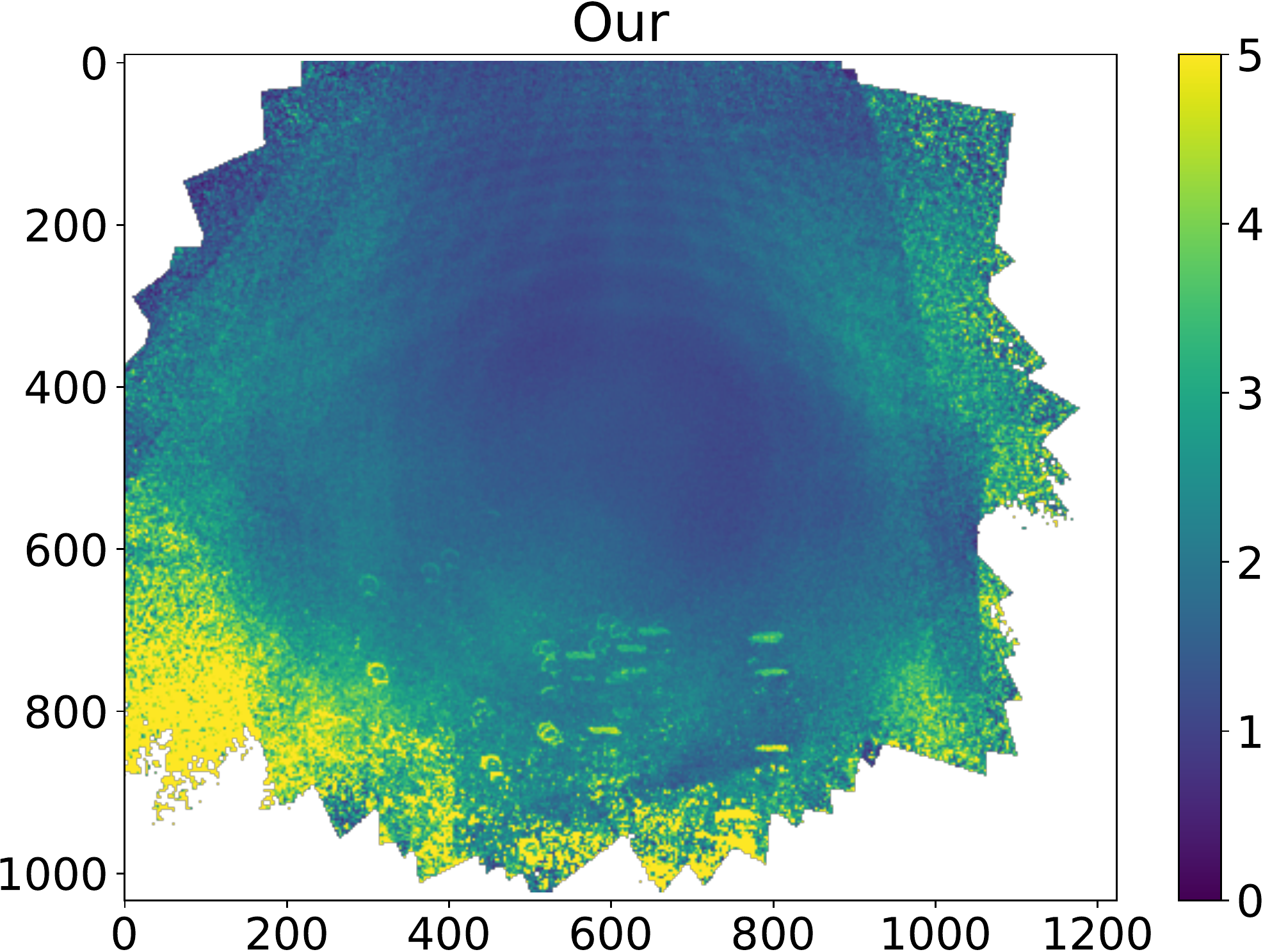}
    \caption{Average per-pixel error in degrees of the produced AoLP for three models. Note that the color bar for the Orthographic model is substantially different to avoid saturation.}
    \label{fig:aolp_comparison}
\end{figure*}

Solving a small linear system for each pixel (Step 4) is the price to pay to
compensate the effects of tilted polarizers. However, this operation can be
used to seamlessly demosaic a DoFP camera. Indeed, in such cameras each pixel
can only observe a single polarizer angle $\alpha_i$ similar to how a pixel
observes a specific color in color cameras with Bayer pattern.
Eq.~\ref{eq:linearsystem} can be slightly adapted to accept groups of $H\times
H$ neighbours of the $j^{th}$ pixel to produce $S_j$. The approach is similar
to \cite{zhang2016image} but can now take into account the effective angles of
the micro polarizers. Moreover, the synthetized images produced in the Step
5 satisfy physical constraints deriving from the orthogonality of the
polarizers. Indeed, it is guaranteed that
$\hat{I}_0+\hat{I}_{90}=\hat{I}_{45}+\hat{I}_{135}$ which was not true in
general for raw images in $\mathcal{I}$ regardless the effect of the tilted
polarizers. This problem was rarely addressed in the literature but can
introduce biases in the computation of Stokes vector since we implicitly give
more importance to some polarizer angles ($0$ and $90$ degrees) with respect to
the other pair. With our approach this problem disappears and corrections like
the one proposed in~\cite{fatima2022one} are no longer required.

\section{Experiments}

In this section we present some experiments to evaluate the ability of the
proposed model to accurately describe how the Stokes vector are imaged by
a projective camera. The two other similar methods available in the literature
are the well known Orthographic model and the recent Perspective Phase Angle
(PPA) proposed by Chen et. al.\cite{chen2022perspective}. As discussed before,
we recall that PPA just relates the AoLP $\phi$ of the light to the surface
normal without giving a unified description of what happens to the whole Stokes
vector. Therefore, comparisons against PPA is limited to specific experiments
discussed in \cref{sec:aolp_recovery,sec:planeorientation}. To reduce
uncertainty due to uncontrollable scene conditions (mixed polarization,
accuracy of the Ground Truth surface normals, $\frac{\pi}{2}$-ambiguity, etc.)
we follow the approach of \cite{chen2022perspective} that uses a glossy planar
plastic board with markers attached to it. In this way, the orientation of the
plane (i.e. its normal in camera reference frame) can be accurately recovered.
We created from scratch a new dataset similar to the PPA dataset provided in
\cite{chen2022perspective} but using a different camera and lenses.
Specifically, we used a FLIR Blackfly camera mounting a $5$ Mpixel Sony
IMX250MZR DoFP sensor with $8$mm lenses. Scene was illuminated with sunlight in
a very overcast day so that the DoLP of incoming light is close to $0$. The
resulting dataset is composed by $360$ images of the glossy plane taken at
different angles and distances. Both PPA and our datasets are used for
evaluation. 

\subsection{AoLP recovery}\label{sec:aolp_recovery}

\begin{table}[]
\begin{center}
\setlength\tabcolsep{4pt} 
\begin{tabular}{|l|cc|cc|}
\cline{2-5}
\multicolumn{1}{c|}{} & \multicolumn{2}{c|}{PPA Dataset} & \multicolumn{2}{c|}{Our Dataset} \\
\cline{2-5} 
\multicolumn{1}{c|}{} & MAE  & RMSE  & MAE  & RMSE  \\ \hline
Ortho      & $13.60 \pm 8.66$ & $16.13$ & $7.66 \pm 9.48$ & $9.50$ \\ \hline
PPA \cite{chen2022perspective} & $1.95 \pm 1.56$ & $2.52$ & $4.89 \pm 6.43$ & $6.50$ \\ \hline
Our       & $\mathbf{1.54} \pm \mathbf{1.39}$ & $\mathbf{2.10}$ & $\mathbf{4.68} \pm \mathbf{6.27}$ & $\mathbf{6.25}$ \\ \hline
\end{tabular}
\caption{MAE and RMSE of the AoLP produced with the three compared models from the GT normal vectors of the plane. Values are in degrees.\vspace{-0.8cm} \label{table:aolpcomparison}}
\end{center}
\end{table}

We started by measuring the agreement between the AoLP $\phi_c$ captured by the
camera and the expected AoLP $\phi_m$ computed by the three models (Ortho, PPA,
and Our). In the Orthographic model $\phi_m=\mbox{atan2}(n_y,n_x)$ (i.e. the
AoLP is equal to the azimuth angle of the surface normal $\vec{n}=(n_x, n_y,
n_z)^T$), in PPA is given by Eq.~6 in \cite{chen2022perspective}, and in our
model is computed like the Orthographic model but with the pre/post-processing
explained in \cref{sec:howtouse}. We manually checked that specular reflection
is dominant in both the PPA and our dataset, so the $\pi$-ambiguity is the only
one that can happen on the AoLP. Therefore, in computing the errors we take the
best between $\phi_m$ and $\phi_m+\pi$.

Results are listed in \cref{table:aolpcomparison}. Both PPA and our model
return a significantly lower error than the Orthographic model. This result,
consistent to what declared in~\cite{chen2022perspective}, should raise the
awareness that ignoring the perspective distortion is probably not reasonable
for accurate Shape-from-polarization. Our model performs better in both the
datasets because it compensates the effect of tilted polarizers. Both the MAE
and RMSE of our model shows an impovement of $\approx25\%$ wrt. PPA. In
\cref{fig:aolp_comparison} we show the per-pixel Mean Absolute Error (MAE) when
computing the AoLP on the whole PPA dataset. Not considering some small
artefacts at the bottom due to specular highlights that saturate some pixels,
we observe a strong radial pattern in the Orthographic MAE (left-most image).
This is expected, since rays corresponding to pixels farther away to the
principal point are more tilted than the ones closer to it. In our model this
effect is strongly attenuated, showing evidence that the perspective distortion
have been effectively compensated. PPA performs better than Ortho, but the
radial pattern is still partially visible.

\subsection{Plane orientation estimation}\label{sec:planeorientation}

\begin{figure}
    \centering
    \includegraphics[scale=0.25]{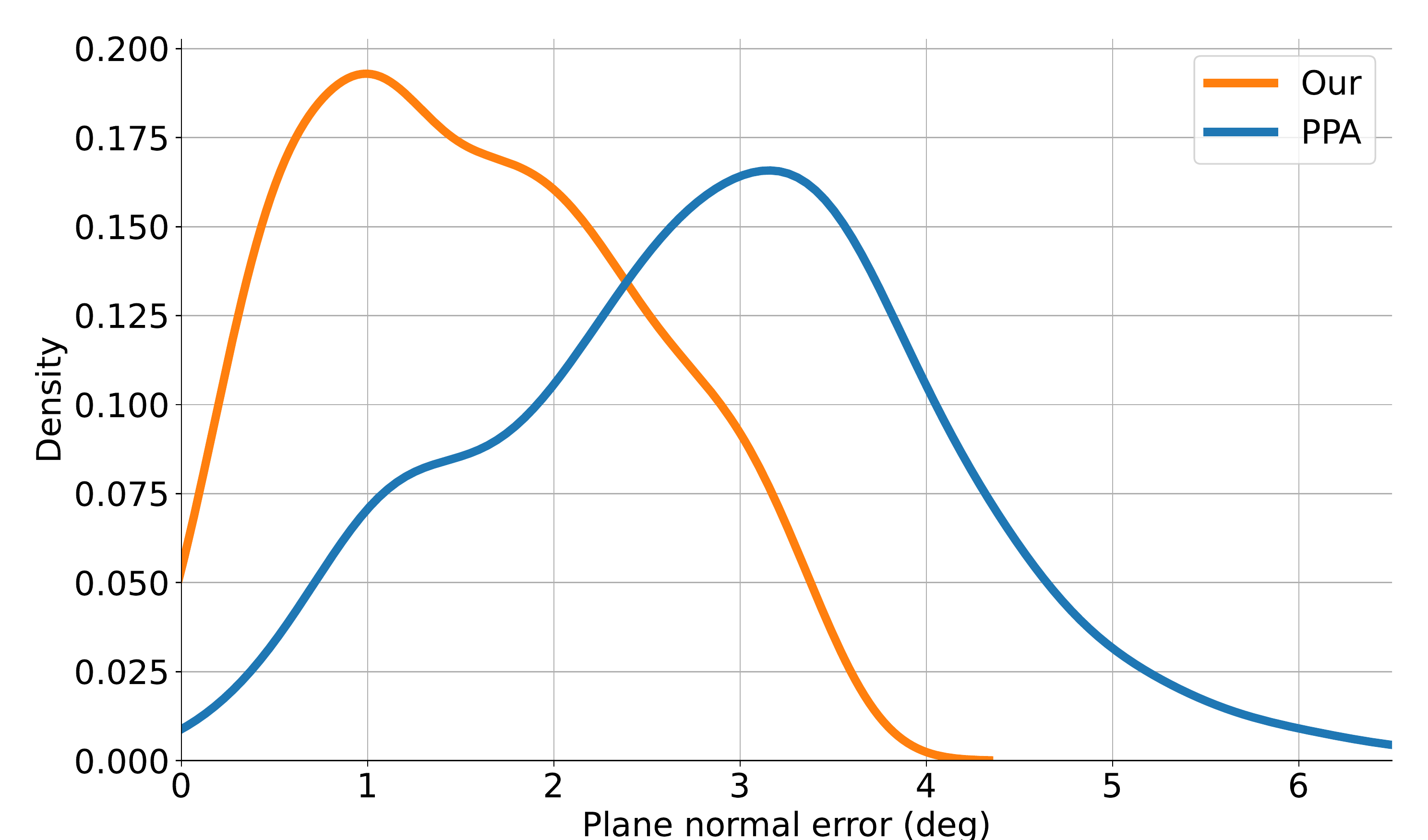}
    \caption{Plane normal error distributions for our method and PPA.}
    \label{fig:plane_orientation_comparison}
\end{figure}

\begin{table*}[]
\begin{center}
\setlength\tabcolsep{14pt}
\begin{tabular}{|l|cc|cc|}
\cline{2-5}
\multicolumn{1}{c|}{} & \multicolumn{2}{c|}{PPA Dataset} & \multicolumn{2}{c|}{Our Dataset} \\
\cline{2-5} 
\multicolumn{1}{c|}{} & MAE  & RMSE  & MAE  & RMSE  \\ \hline
Orthographic   & $13.441 \pm 6.653$  & $14.997$ & $10.546 \pm 5.467$ & $11.879$ \\ \hline
Our    & $1.923 \pm 1.343$ & $2.346$ & $4.373 \pm 2.935$ & $5.267$ \\ \hline
Smith       & $28.494 \pm 12.379$ & $31.067$ & $34.848 \pm 5.477$ & $35.275$ \\ \hline
Smith corrected    & $27.396 \pm 12.489$ & $31.021$ & $33.870 \pm 6.399$ & $34.470$ \\ \hline
SfPW CNN      & $38.305 \pm 22.915$ & $44.637$ & $22.819 \pm 15.649$ & $27.670$ \\ \hline
\end{tabular}
\caption{Comparison of normal estimation errors on PPA and our plane datasets
between different SfP methods. Projective (Our) and Orthographic methods use an
oracle to solve all the ambiguities. Errors are in degrees.}
\label{table:normal_plane_comparison}
\end{center}
\end{table*}

Since in the Orthographic model $\phi$ is the azimuth angle of $\vec{n}$, the
linear constraint
\begin{equation} \begin{pmatrix} \sin \phi & \cos \phi & 0\end{pmatrix} \cdot
\vec{n} = 0 \end{equation}
\noindent is usually considered in photo-polarimetric stereo approaches or
iso-depth contour tracing. If we know that at least $K>3$ pixels observe the
same plane, we can use such constraint to recover the plane normal $\vec{n}$
(in camera reference frame) from the (corrected) $\phi_j$ observed at each
pixel by solving:
\begin{equation} \begin{pmatrix} \begin{pmatrix} \sin \phi_1 & \cos \phi_1
& 0 \end{pmatrix} R_1^T \\ \begin{pmatrix} \sin \phi_2 & \cos \phi_2
& 0 \end{pmatrix} R_2^T \\ \vdots \\ \begin{pmatrix} \sin \phi_K & \cos \phi_K
& 0 \end{pmatrix} R_K^T \\ \end{pmatrix} \vec{n} = \vec{0} \end{equation}
\noindent as a Linear Least-Squares problem. Note that the system is
under-determined in the Orthographic model because all the rays are parallel
($R_1, \ldots R_k$ are identities) and all the pixels would observe the same
$\phi$. In the PPA model, instead, a similar costraint is provided (See Eq.
7 in \cite{chen2022perspective}).

In \cref{fig:plane_orientation_comparison} we plotted the the estimated plane
normal error distribution against the Ground Truth data in the PPA dataset. The
Ortho model is not present in the plot for the reasons discussed before. Also
in this case, our mean absolute error is lower ($1.57^\circ$ vs. $2.89^\circ$)
and with less variability. This reflects a better estimation of the AoLP due to
the correction applied by the tilted polarizer model.

\begin{figure*}
    \centering
    \includegraphics[width=0.33\linewidth,trim={0 60 0 0},clip]{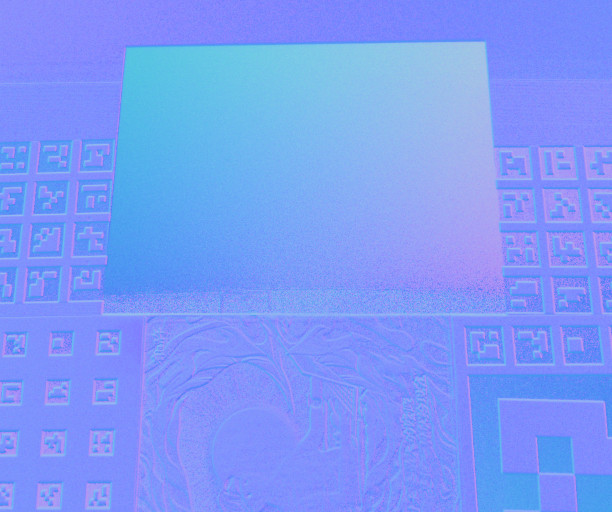}
    \includegraphics[width=0.33\linewidth,trim={0 60 0 0},clip]{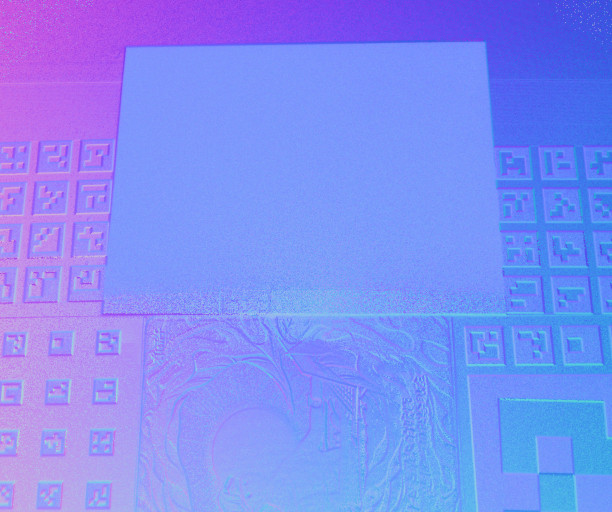}
    \includegraphics[width=0.33\linewidth,trim={0 60 0 0},clip]{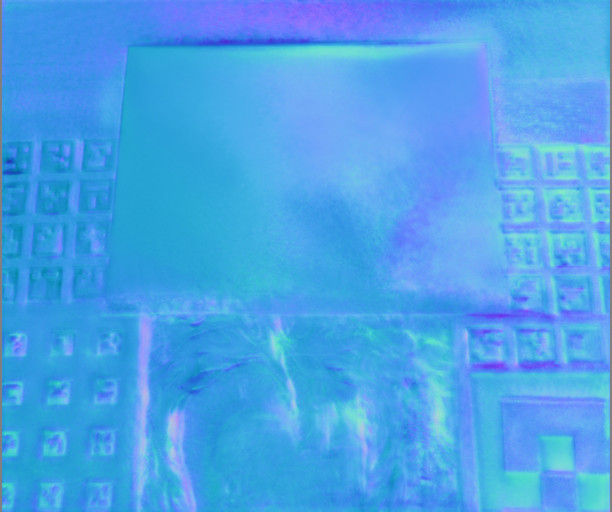}
    \caption{Qualitative comparison of the estimated normal field with methods
Ortho (left), Our (center), and SfPW (right) on an instance of the PPA dataset.
Errors are computed only inside the plane area but we kept the rest of the
scene for visualization purposes. Note how a radial pattern is visible in the
Ortho model since the distortion due to the ray orientation is not accounted
for. SfPw performs poorly because it has been trained on a much noisier dataset
and apparently is not able to generalize to simpler scenarios.}
    \label{fig:qualitative_comparison}
\end{figure*}

\subsection{Normal estimation}

Our work is not a SfP method, so evaluating its validity based on how well
a normal field is reconstructed has to be done with care. We recall that the
azimuth angle of $\vec{n}$ is equal to the AoLP $\phi$ up to a (unavoidable)
$\pi$-ambiguity, and an additional $\frac{\pi}{2}$-ambiguity depending if
specular or diffuse reflection dominates. Since in the plane dataset the
specular reflection is dominant, we used the following function to relate the
DoLP $\rho$ with zenith angle $\theta$ and the index of refraction $n$ :
\begin{equation}\label{eq:dolpspecular}
    \rho(n,\theta)=a \frac{2 \sin^2 \theta \cos \theta \sqrt{n^2 - \sin^2 \theta}}{n^2-\sin^2\theta-n^2\sin^2\theta +2 \sin^4\theta}
\end{equation}
\noindent where $a$ is a scale parameter we added to the original formulation
shown in \cite{atkinson2006recovery} to account for a possible diffuse
component of reflection and other nonlinear contribution like the Umov's effect
\cite{kupinski2019angle,umow1905chromatische}.

To show the validity of our approach, the idea is to test how well our model
allows the recovery of surface normals in ideal conditions, i.e. with an oracle
that removes the ambiguities and provides a close estimate of the unknown
parameters $n,a$ in Eq.~\ref{eq:dolpspecular}. Since we know the Ground Truth
normals of both the datasets, we manually estimated the best value of the two
parameters and computed the errors considering the minimum error among all the
possible ambiguities. We compare the resulting normal field against: (i) the
\emph{orthographic} model using the same oracle as our method (i.e. estimated
function $\rho$ and optimal disambiguator); (ii) \emph{Smith} et al.
\cite{smith2018height}; (iii) \emph{Smith corrected} with our pre/post
processing to account for projective camera, and (iv) SfPW \cite{lei2022shape},
a Deep CNN with a multi-head self-attention module and viewing encoding ``to
account for non-orthographic projection in scene-level". We used the original
author implementation pre-trained on their SPW Dataset.

We listed the MAE and RMSE of the estimated normals in
\cref{table:normal_plane_comparison}. Normals recovered with our projective
model are significantly more accurate than the others. This is not surprising,
since we assume to have an oracle that resolves all the unknowns we usually
face when doing SfP. The important thing to notice here is the improvement
against the orthographic model that uses the same oracle. In other words,
properly accounting for non orthographic projection is crucial to reduce the
error, no matter how sophisticated is the method applied to solve the
ambiguities.
 Moreover, the \emph{Smith corrected} performs better than the original one
because can now accounts for non-orthographic cameras. We stress that this
correction was performed as pre/post processing without modifying any other
part of the method. Finally, SfPW performs well but not better than the
``simple" Orthographic model. Considering that SfPW can take into account
non-orthographic cameras, we expected a lower error in this experiment. In
general, SfP is difficult to solve without posing additional strong priors to
the reconstructed scene, and learning-based method can only try to resolve the
ambiguities based on what observed on the training set, with the risk of
overfitting specific 3D structures. In \cref{fig:qualitative_comparison} we
show a qualitative example of the obtained normal field.

\subsection{Other datasets}

We also tested our method on a dataset of general objects presented in
\cite{ba2020deep}. In this case it is difficult to provide an oracle because it
contains scenes with different kind (diffuse/specular) and degrees of
polarization. So, we limit our test to the observation of how the pre/post
processing can improve an existing SfP method. Similarly to what we did with
planes dataset, we compare \emph{Smith}, \emph{Smith corrected}, \emph{SfPW}
and \emph{DeepSfP}\cite{ba2020deep}. The latter method is a Deep model assuming
an orthographic camera, so potentially is a valid candidate to be corrected
with our model. However, at the time of writing, the code and pre-trained
weights were not publicly available so we report here the results shown in the
original paper. \cref{table:deepsfp_comparison} show the MAE of the estimated
normals in the whole dataset. SfPW is the best performing method, followed by
DeepSfP. It is interesting to observe that, also in this case, Smith corrected
with our model obtains a significant improvement.

\begin{table}[b]
\begin{center}
\begin{tabular}{|l|c|}
\cline{2-2} 
\multicolumn{1}{c|}{} & MAE (deg) \\ \hline
Smith           & $45.39$    \\ \hline
Smith corrected & $32.52$   \\ \hline
SfPW CNN        & $14.68$   \\ \hline
DeepSfP         & $18.52$    \\ \hline
\end{tabular}
\caption{Normal estimation errors on DeepSfP dataset}
\label{table:deepsfp_comparison}
\end{center}
\end{table}

\section{Conclusions}

We presented a model to describe how a projective camera captures the light
polarization state. Differently than the empirical PPA
model~\cite{chen2022perspective}, our formulation derives by the optical
properties of the tilted polarizer and the algebra of Mueller matrices.  We
consider it as a unifying model, equally valid to DoFP and ToFP cameras (i.e.
with polarizers above the sensor or in front of the lenses) and consistent with
conventions used by polarization-aware renderers. It allows to embed
demosaicing of DoFP cameras directly in the Stokes estimation and, finally, can
be implemented by means of a pre-processing of acquired raw data and
post-processing of the reconstructed surface normals. In the experimental
section we observed a good agreement between the obtained Stokes vector (in
terms of AoLP and DoLP linked to the surface normals) and the expected one
deriving by the geometry of some controlled planar scenes. For the future, we
aim to generalize our model to include lens distortion, vignetting on acquired
DoLP due to the Umov's effect, and noise deriving by cross-talking between
pixels.

{\small
\bibliographystyle{ieee_fullname}
\bibliography{egbib}
}

\end{document}